\documentclass[11pt]{article}

\usepackage[final]{acl}

\usepackage{times}
\usepackage{latexsym}

\usepackage[T1]{fontenc}

\usepackage[utf8]{inputenc}

\usepackage{microtype}

\usepackage{inconsolata}

\usepackage{graphicx}

\usepackage{multirow}
\usepackage{booktabs}
\usepackage[most]{tcolorbox}
\tcbset{colframe=gray, colback=white, boxrule=0.5pt, boxsep=8pt}
\usepackage{color, soul}
\usepackage[dvipsnames]{xcolor}    
\usepackage{here}
\usepackage{colortbl}
\usepackage{array}
\usepackage{here}
\usepackage{tabularx}
\usepackage{subcaption}

\definecolor{lightgray}{rgb}{0.88, 0.88, 0.88}

\newcommand{\kosuke}[1]{\textcolor{black}{#1}}

\title{Improving Mathematical Reasoning Capabilities in Large Language Models via Reasoning Process Error Classification}

\author{Runa Yoshida, Kosuke Nishida, Kyosuke Nishida \\
    Human Informatics Labs., NTT, Inc.\\
  \texttt{\{runa.yoshida, kosuke.nishida, kyosuke.nishida\}@ntt.com}\\ }

\begin{document}
\maketitle
\begin{abstract}
    The reasoning ability of large language models (LLMs) is a critical factor for practical LLM-based applications.
    To investigate the current reasoning capability of LLMs, 
    we clarify the types of errors that arise in LLMs' reasoning processes on mathematical datasets.
    We focus on problems where LLMs produce an incorrect answer.
    We define errors in the reasoning process as reasoning errors and manually analyze the features of reasoning errors.
    We defined and classified 21 error classes and identified the frequently occurring classes among them.
    Beyond qualitative evaluation, we leverage the evaluation results to improve the reasoning capability.
    We designed a prompt that explicitly focuses on eight error classes.
    The experiments demonstrate that this prompt effectively improves reasoning performance.
    Furthermore, the results suggest that the frequent reasoning errors identified in this paper are common across LLMs of comparable scale.

\end{abstract}

\section{Introduction}

    Large language models (LLMs) demonstrate high performance across a wide range of tasks, which has led to growing interest in their practical applications.
    The reasoning capabilities of LLMs are critically important for practical applications such as scientific computing, which require the ability to derive conclusions that satisfy multiple conditions and to decompose complex problems into simpler subproblems.
    
    Mathematical reasoning is a task that requires strict logical consistency and computational accuracy; consequently, mathematics benchmarks have been adopted to evaluate the reasoning capabilities of LLMs~\citep{cobbe2021gsm8k, hendrycksmath2021}.
    However, the model’s reasoning capabilities remain limited:
    \citet{boye2025largelanguagemodelsmathematical} and \citet{mirzadeh2025gsmsymbolic} show that models struggle with complex questions and with problems requiring multi-step deduction or real-world knowledge.
    
    In this paper, we investigate the current state of LLM reasoning capabilities through a thorough analysis focusing on why LLMs arrive at incorrect answers in the mathematical domain.
    Specifically, we analyze errors in the reasoning process that lead to incorrect answers, which we refer to as reasoning errors.
    For reasoning errors observed in the analyzed problems, we manually grouped similar errors into the same class and identified \textit{reasoning error classes}. 
    We collected reasoning errors from 583 questions in the MATH benchmark~\citep{hendrycksmath2021} test set, on which the Llama-3.3-70B-Instruct\footnote{https://huggingface.co/meta-llama/Llama-3.3-70B-Instruct}~\citep{llama3modelcard} model failed to arrive at the correct answer.
    
    As a result of our analysis, we identified 21 error classes.
    The classification results revealed challenges of current LLMs:
    (i) logical reasoning, 
    (ii) understanding the context of problem conditions and settings, 
    (iii) calculations and algebraic manipulations,
    (iv) insufficient consideration of all the conditions in the problem statement, and
    (v) insufficient consideration of the prerequisites of the introduced theorems.

    Building upon the evaluation results, we utilize a training-free approach to reduce errors in frequent classes.
    We design a prompt that explicitly pays attention to frequently occurring reasoning error classes.
    Our approach is consistent with the findings of \citet{tyen-etal-2024-llms} that the primary limitation of LLM reasoning lies in detecting errors rather than correcting them.
    
    Experimental results show that the prompt designed with consideration of reasoning error classes enhances performance on mathematical datasets with statistical significance.
    Furthermore, since the proposed method shows consistent gains among three LLM families, this suggests that the reasoning errors identified in this paper are common challenges for LLMs. 
    This demonstrates that systematically classifying and analyzing reasoning errors is not only useful for model evaluation but also serves as a valuable guideline for the practical application of LLMs.

    In summary, our contributions are:
    \begin{itemize}
        \item Through an analysis of reasoning errors, we identified the tendencies of reasoning errors exhibited by LLMs. \kosuke{These classes will provide a novel framework for evaluating LLMs' capability in the mathematical domain.}
        \item We show that a prompt based on the analysis of reasoning errors can effectively improve the reasoning capabilities of LLMs.
        \item The reasoning errors identified in this paper suggest that they are common challenges encountered by LLMs of comparable scale.
    \end{itemize}

\section{Related Work}

    \paragraph{Classification of reasoning errors. }
    Classification of reasoning errors in LLMs is an important topic~\citep{Solving_Quantitative_Reasoning, 
    bubeck2023sparksartificialgeneralintelligence, 
    golovneva2023roscoe, seßler2024benchmarkinglargelanguagemodels, 
    boye2025largelanguagemodelsmathematical}.    
    Among these studies, our work conducts evaluations through error classification and leverages the results to improve LLMs' ability.
    Similar to our motivation, \citet{yu-etal-2025-self-error} and \citet{pan-etal-2025-lemma} aim to leverage evaluation results to improve LLMs.
    \kosuke{These studies} classified the reasoning errors exhibited by models on a mathematics dataset into eight or twelve classes using an LLM and demonstrated that mathematical reasoning performance can be improved by creating training data corresponding to each class.
    On the other hand, 
    the error classes they used consist of both domain-specific categories (e.g., Geometric Errors) and coarse-grained categories (e.g., Logical and Reasoning Errors), as their primary objective was to construct training data.
    This study introduces new classes that generalize and expand their classes, presenting more detailed error analysis.
    
    \kosuke{Moreover, \citet{yin-etal-2025-error} proposed a retrieval-based method. In advance, the method collects problem and incorrect solution pairs, and generated error analysis for each pair. 
    For inference, it dynamically retrieves error analyses relevant to the input problem and incorporates them into the prompt to encourage the model to pay attention to these errors.}
    Our study identifies reasoning error tendencies in a single model and shows that a fixed, lightweight prompt derived from these tendencies improves mathematical reasoning performance across different model families.
    
    Furthermore, several studies showed that even high-performing LLMs exhibit varying error detection and classification performance across datasets and still perform inadequately on challenging problems~\citep{ijcai2024p381, singh-etal-2025-exposing, yin-etal-2025-error}.
    Therefore, we manually classified the reasoning errors rather than using LLM-based classification.

    \paragraph{Enhancing reasoning ability via prompt design.}
    Currently, various Chain-of-Thought (CoT) prompting techniques have been extensively investigated, and these approaches have been shown to enhance the problem-solving capabilities of LLMs~\citep{WeiCoT2022, Self-Consistency}.
    For example, appending the phrase “Let's think step by step” to the end of a question has been reported to prompt LLMs to generate explicit step-by-step reasoning processes, which in turn reduces reasoning errors~\citep{NEURIPS2022_8bb0d291}.
    \citet{li-etal-2024-evaluating-mathematical} use GPT-4~\citep{openai2024gpt4technicalreport} to classify errors into nine error types 
    and show that explicitly including the assigned error class labels in the prompt improves large-scale LLMs’ error correction ability.
    However, the gains were limited for open-source models such as Llama-2~\citep{touvron2023llama2openfoundation}, which suggests that error class labels alone may not provide sufficient guidance for models that struggle to estimate the underlying cause of an error. 
    Following these studies, we transform frequently observed reasoning error classes into verification questions and incorporate them into the prompt to reduce such errors in LLMs under zero-shot prompting.

\section{Analysis of Reasoning Errors}
\label{sec:analysis}
    We describe our analysis procedure. 
    It consists of methods for \kosuke{the selection of target } problems, the approach to classifying reasoning errors, and the classification results. \kosuke{Then, we present a discussion based on the analysis.}

    \begin{table}[t]
            \centering
            \small
            \tabcolsep3pt
            \begin{tabular}{l|r|rrrrr|r}
                \toprule
                & MATH &\multicolumn{6}{c}{\# Analyzed } \\
                \cmidrule(lr){2-2}
                \cmidrule(lr){3-8}
                & \multicolumn{1}{c|}{Total} &{\tiny Lev.} 1 & 2& 3& 4 &  5 & Total  \\
                \midrule
                Algebra   & 1,187  &  0&  3&  5&  8& 25& 41 \\
                Counting and & \multirow{2}{*}{474} &  \multirow{2}{*}{1}& \multirow{2}{*}{10}&  \multirow{2}{*}{9}& \multirow{2}{*}{16}& \multirow{2}{*}{37}&  \multirow{2}{*}{73} \\
                \qquad Probability &&&&&&&\\ 
                Geometry   & 479  &  8& 16& 14& 36& 50& 124 \\
                Intermediate & \multirow{2}{*}{903} &  \multirow{2}{*}{1}&  \multirow{2}{*}{9}& \multirow{2}{*}{23}& \multirow{2}{*}{31}& \multirow{2}{*}{68}& \multirow{2}{*}{132}  \\
                \qquad Algebra &&&&&&&\\ 
                Number Theory    & 540 &  0&  7& 15& 22& 40&  84 \\
                Prealgebra & 871 &  2&  7& 10& 17& 40&  76  \\
                Precalculus & 564&  2&  7& 10& 12& 22&  53  \\
                \midrule
                Total      & 5,000 & 14& 59& 86&142&282& 583  \\
                \bottomrule
            \end{tabular}
            \caption{Total number of problems in the MATH test set and the number of \kosuke{the target problems.}} 
            \label{tab:target}
        \end{table}

    \subsection{Selection of \kosuke{Target} Problems for Analysis}
\begin{figure}[t]
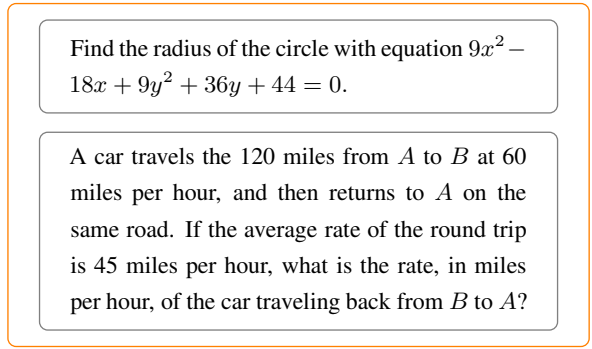

            \centering
            \begin{tcolorbox}[colframe=orange, colback=white, boxrule=0.5pt, boxsep=0pt]
            \begin{tcolorbox}[colframe=gray, colback=white, boxrule=0.5pt, boxsep=0pt]
                {\small 
                Find the radius of the circle with equation $9x^2-18x+9y^2+36y+44=0.$
                }
            \end{tcolorbox}
            \begin{tcolorbox}[colframe=gray, colback=white, boxrule=0.5pt, boxsep=0pt]
                {\small 
                A car travels the 120 miles from $A$ to $B$ at 60 miles per hour, and then returns to $A$ on the same road. If the average rate of the round trip is 45 miles per hour, what is the rate, in miles per hour, of the car traveling back from $B$ to $A$?
                }
            \end{tcolorbox}
            \end{tcolorbox}
            \caption{Examples of level 5 algebra problems.}
            \label{fig:Algebra_level5_example}
        \end{figure}

        \paragraph{Problem source.}
        For our analysis, we adopted the MATH benchmark test set.
        The MATH dataset includes problems ranging from secondary education to early undergraduate level and is divided into five difficulty levels.
        It consists of seven domains: algebra, counting and probability, geometry, intermediate algebra, number theory, prealgebra, and precalculus.
        Figure \ref{fig:Algebra_level5_example} shows examples of level 5 problems in the algebra domain.
        For each problem, both the reasoning process and the final answer are provided as annotations.
        \citeposs{seßler2024benchmarkinglargelanguagemodels} said that this dataset contains numerous high-difficulty problems, which are challenging for correct reasoning even by the Llama-3-70B model\footnote{https://huggingface.co/meta-llama/Meta-Llama-3-70B}~\citep{llama3modelcard}. 
        Consequently, we consider it appropriate for our analysis.

        \begin{table*}[p]
            \centering
            \tabcolsep3pt
            \begin{tabularx}{\linewidth}{lX|r}
                \toprule
                \rowcolor{lightgray}%
                (I) & Problem Understanding Errors &  185\\
                (I-1) & Misreading of Information Contained in Figures, Tables, Charts, and Graphs                & 69\\
                (I-2) & Ignoring Conditions in the Problem Statement   & 53\\
                (I-3) & Introducing Conditions not in the Problem Statement       & 31\\
                (I-4) & Misinterpretation of Conditions in the Problem Statement              & 27\\
                & {\small The model misinterprets a given condition and performs reasoning based on that incorrect interpretation.} & \\
                (I-5) & Misinterpretation of Range-, Quantity-, and Comparison-Related Conditions                    &  5\\
                \midrule
                \rowcolor{lightgray}
                (II) & Mathematical Concept Knowledge Errors    &  \phantom{0}32\\
                (II-1) & Misremembering Theorems and Formulas        & 19\\
                (II-2) & Confusion between Concepts with Different Definitions or Properties          & 13\\
                \midrule
                \rowcolor{lightgray}
                (III) & Arithmetic and Algebraic Manipulation Errors                   &  135\\
                (III-1) & Incorrect Numerical Calculations             & 61\\
                (III-2) & Algebraic Misoperations                & 39\\
                (III-3) & Miscounting the Number of Elements in a Set or Range                     & 31\\
                (III-4) & Misinterpretation of Numerical Magnitude and Order  &  4\\
                \midrule
                \rowcolor{lightgray}
                (IV) & Logical and Reasoning Errors               &  230 \\
                (IV-1) & Application of Inappropriate Theorem        & 77\\
                &{\small The model applies a theorem without satisfying its prerequisites or incorrectly handles its conclusion in a mathematical reasoning step.}\\
                (IV-2) & Incorrect Logical Implication & 57\\ 
                &{\small Even though the current statement is true, the model presents an argument that contains a logical error, leading to a conclusion that does not logically follow.}\\
                (IV-3) & Incorrect Final Answer Selection      & 38\\
                & {\small The model incorrectly selects a final answer from the derived answer candidates.}& \\
                (IV-4) & Inappropriate Strategy Selection                & 18\\
                & {\small The model selects inappropriate strategies or plans.}& \\
                (IV-5) & Misapplication of Formulas and Theorems    & 14\\
                & {\small The model uses incorrect substituted values or corresponding relationships, even though the conditions or premises are satisfied.} &\\
                (IV-6) & Incorrect Reasoning of Conditions in the Problem Statement & 12\\
                & {\small The model misprocesses the conditions in the problem statement, leading to the selection of inappropriate solution methods or theorems.} & \\
                (IV-7) & Incorrect Formulation of Regularity and Repetitive Structures &  9\\
                (IV-8) & Inappropriate Processing                  &  5\\
                & {\small The model introduces inappropriate operations, such as rounding, leading to subsequent calculations or incorrect reasoning.} & \\
                \midrule
                \rowcolor{lightgray}
                (V) & Others                      &   \phantom{0}52 \\
                (V-1) & Transcription Mistakes                  & 37\\
                & {\small The model outputs an incorrect value as the final answer despite having derived the correct conclusion.} & \\
                (V-2) & Correct or Non-error                      & 15\\
                & {\small This class covers cases where the dataset annotation is incorrect, or the model's answer is actually correct (e.g., whether a unit is included).} & \\
                \bottomrule
            \end{tabularx}
            \caption{
            Defined reasoning error classes and the number of classified errors.
            \kosuke{For each coarse-grained class, the count is the sum of the counts of its finer-grained classes.}
            Some problems fall into multiple classes of reasoning errors.
            }
            \label{tab:category}
        \end{table*}
        
        \paragraph{Generation of reasoning processes and answers.}
        We generated reasoning processes and answers using  Llama-3.3-70B-Instruct.
        We adopted a 70B-scale model based on observations in \citet{seßler2024benchmarkinglargelanguagemodels} that 8B-scale models have limited arithmetic reasoning capabilities.
        For each problem, CoT reasoning is performed with a sequence length fixed at 1,024, and answers are generated using greedy decoding in a zero-shot setting.
        We excluded problems where the model could not produce a final answer due to the sequence length constraint.
        
        \paragraph{Judgment of correctness.}
        The target of the analysis is problems for which the model provided incorrect answers.
        To determine whether the model answered correctly,
        we used Exact Match (EM) between the model's final answer and the final answer annotated in the dataset.
        However, EM-based evaluation may incorrectly classify answers as incorrect even when they are mathematically equivalent.
        For example, although $-\sqrt{2} + 1$ and $1 - \sqrt{2}$ are mathematically equivalent, EM-based evaluation considers them incorrect.
        To address this issue, we used gpt-oss-120b\footnote{https://huggingface.co/openai/gpt-oss-120b}~\citep{openai2025gptoss120bgptoss20bmodel} to determine whether the two answers were equivalent. 
        When answers were judged to be equivalent, we excluded their problem from the analysis after manual verification.
        Table~\ref{tab:target} shows the number of problems selected for analysis.

        \paragraph{Implementation.}
        We used the evaluation script provided by DeepSeek-MATH\footnote{https://github.com/deepseek-ai/DeepSeek-Math/tree/main/evaluation}~\citep{deepseek-math}.
        The evaluation procedure, including prompt configuration, followed the same script.
        
    \subsection{Definition and Classification of Reasoning Error Classes}
        We defined classes of reasoning errors that occurred in the target problems and manually classified them.
        We used error classes identified in prior studies~\citep{yu-etal-2025-self-error, pan-etal-2025-lemma, boye2025largelanguagemodelsmathematical} as first-level categories.
        We further introduced finer-grained subcategories and defined them as the error classes in this study.
        First, we carefully examined the reasoning process in \kosuke{the} target problems that did not reach the correct answer, and categorized reasoning errors by coarse granularity as follows:
        (I) Problem Understanding Errors, 
        (II) Mathematical Concept Knowledge Errors, 
        (III) Arithmetic and Algebraic Manipulation Errors, 
        (IV) Logical and Reasoning Errors, and
        (V) Others.
        We excluded domain-dependent reasoning error classes, such as Geometric Errors, provided by prior work.
        Next, we compared reasoning errors within each coarse-grained category and defined finer-grained reasoning error classes \kosuke{by} grouping together similar errors.
        As a result, we defined a total of 21 finer-grained reasoning error classes.

    \begin{figure*}[t]
        \centering
        \begin{tcolorbox}
            {\color{blue}\{problem\}} \newline
            {\color{blue}Please reason step by step, and put your final answer within \textbackslash\textbackslash boxed\{\}.}
            When answering questions and explaining your reasoning process, please pay particular attention to the following perspectives at each step:\\
            - Have all conditions within the text been extracted? $\textcolor{gray}{_{(\text{I}-2)}}$\\
            - Have you introduced any implicit assumptions not stated in the problem? $\textcolor{gray}{_{(\text{I}-3, \ 4,\ \text{IV}-1)}}$ \\
            - Have the prerequisites for the theorems and formulas used been verified? 
            $\textcolor{gray}{_{(\text{IV}-1,\ 2)}}$\\
            - Does it satisfy all the conditions and prerequisites of the problem statement? $\textcolor{gray}{_{(\text{IV}-1,\ 3, \ 6)}}$ \\
            - Is it clear what each step does and why the operation is valid? $\textcolor{gray}{_{(\text{IV}-1, \ 2, \ 3,\ 5)}}$ \\
            Do not output confirmation for items that have already been verified. 
        \end{tcolorbox}
        \caption{
        The \kosuke{proposed} 
        prompt that explicitly considers \kosuke{frequent} reasoning error classes.
        The \textcolor{blue}{blue} text denotes Default Prompt.
        The \kosuke{labels} 
        in \kosuke{the} parentheses for each instruction indicates the reasoning error class considered.
        The \kosuke{labels} 
        of the reasoning error classes correspond to \kosuke{those in} Table~\ref{tab:category}:
        (I-2) Ignoring Conditions in the Problem Statement, 
        (I-3) Introducing Conditions not in the Problem Statement, 
        (I-4) Misinterpretation of Conditions in the Problem Statement, 
        (IV-1) Application of Inappropriate Theorem,  
        (IV-2) Incorrect Logical Implication, 
        (IV-3) Incorrect Final Answer Selection, 
        (IV-5) Misapplication of Formulas and Theorems, and
        (IV-6) Incorrect Reasoning of Conditions in the Problem Statement.
        }
        \label{fig:prompt_gen}
    \end{figure*}

    \subsection{Discussions}
        Table~\ref{tab:category} shows the defined reasoning error classes and the number of the problems corresponding to each class.
        Some problems fall into multiple classes of reasoning errors.
        The classification results revealed that (IV) Logical and Reasoning Errors, (I) Problem Understanding Errors, and (III) Arithmetic and Algebraic Manipulation Errors occurred frequently.
        
        Among the finer-grained classes, (IV-1) Application of Inappropriate Theorem was the most frequent.
        This indicates that current LLMs struggle to consider all the conditions in the problem statement and the prerequisites of the introduced theorems.
        Other frequent classes are also considered to be related to this factor:
        (I-2) Ignoring Conditions in the Problem Statement, 
        (I-3) Introducing Conditions not in the Problem Statement, 
        (I-4) Misinterpretation of Conditions in the Problem Statement, 
        and (IV-6) Incorrect Reasoning of Conditions in the Problem Statement.
        \kosuke{That is, we found that a major challenge for current LLMs lies in their ability to accurately extract and interpret conditions from problem statements, and to determine whether these conditions align with the prerequisites of the theorems introduced during their reasoning process.}

        \kosuke{(I-1) Misreading of Information Contained in Figures, Tables, Charts, and Graphs is the second most frequent class. This suggests that, while LLMs demonstrate strong capabilities in mathematical reasoning, the textual understanding of mathematical problems remains challenging. As discussed above, accurate understanding of problem statements plays an essential role in reasoning. Accordingly, the use and evaluation of vision-language models~\citep{llava} is a promising direction. }

        \kosuke{We observe that approximately 6.3\% of the reasoning processes exhibit (V-1) Transcription Mistakes, in which the model successfully derives the correct conclusion but fails to accurately transfer it to the final answer. Although such errors are inherently simple, their presence even in current LLMs suggests that improving reasoning ability requires addressing not only complex reasoning failures but also these seemingly minor inconsistencies.}

        (III) Arithmetic and Algebraic Manipulation Errors account for 23.2\% of all reasoning errors. 
        In particular, 61 cases involve (III-1) Incorrect Numerical Calculations, while 39 involve (III-2) Algebraic Misoperations. 
        These findings suggest that LLMs with tens of billions of parameters can still struggle to accurately execute individual calculations and algebraic manipulations, even when they identify an appropriate mathematical solution strategy.
        Therefore, mechanisms that explicitly verify numerical calculations and algebraic transformations step by step, as well as evaluations of reasoning supported by external tools such as calculators and computer algebra systems, represent promising directions.

        Our analysis revealed the infrequency of (II) Mathematical Concept Knowledge Errors.
        This result is consistent with the observation that \kosuke{LLMs} 
        successfully utilize knowledge but still require significant improvements in reasoning capabilities~\citep{jin-etal-2025-disentangling-memory}.

\section{Proposed Method}
\label{sec:method}

    We propose a prompt that explicitly considers the classification results of reasoning errors. 
    \citet{tyen-etal-2024-llms} argue that a primary limitation of LLMs' reasoning capabilities stems from their limited ability to detect errors.
    Therefore, explicitly directing attention to frequently occurring errors is expected to compensate for the ability to detect errors and contribute to improvements in reasoning performance.

    \kosuke{Specifically, we designed a prompt to guide the model to focus on two groups of frequent reasoning error classes. The first group consists of the classes related to understanding of the problem statements: (I-2), (I-3), and (I-4). The second group consists of the classes regarding the logical errors: (IV-1), (IV-2), (IV-3), (IV-5), and (IV-6). Based on these fine-grained error classes, our prompt includes five questions that verify the reasoning process with respect to the eight error classes.}
    Figure~\ref{fig:prompt_gen} shows the proposed prompt.

\section{Experiment and Result}
\label{sec:experiment}


    In this section, we address three research questions:
    \begin{description}
        \item[RQ1] Is explicitly directing attention to considering reasoning errors effective?
        \item[RQ2] Is fine-grained reasoning error classification effective?
        \item[RQ3] Are frequent reasoning errors model-specific?
    \end{description}

    \subsection{Experimental Settings}
        \paragraph{Baseline.}
        We adopt the zero-shot standard prompt of DeepSeek-MATH (hereafter referred to as Default Prompt; see the \textcolor{blue}{blue} text in Figure~\ref{fig:prompt_gen}).
        We also adopt a prompt that appends the following instruction to the end of Default Prompt (hereafter referred to as Pay Attention Prompt):
        \begin{figure}[H]
            \centering
            \begin{tcolorbox}[colframe=gray, colback=white, boxrule=0.5pt, boxsep=2pt]
                Please pay particular attention when answering questions and explaining your reasoning process.
            \end{tcolorbox}
        \end{figure}
        \noindent
        We adopted Pay Attention Prompt as a baseline to clearly distinguish whether the performance improvement of the proposed method arises from merely directing attention to the reasoning process or from explicitly considering the reasoning error classes.
        
        We also adopt a prompt that appends the following instruction to the end of Default Prompt (Coarse Attention Prompt):
        \begin{figure}[H]
            \centering
            \begin{tcolorbox}[colframe=gray, colback=white, boxrule=0.5pt, boxsep=2pt]
                When answering questions and explaining your reasoning process, please pay particular attention to the following perspectives at each step:\\
                – Is each reasoning step logically valid? $\textcolor{gray}{_{(\text{IV}) \text{ Logical and Reasoning Errors}}}$\\
                – Have you accurately understood the problem statement and its conditions? $\textcolor{gray}{_{(\text{I}) \text{ Problem Understanding Errors}}}$
            \end{tcolorbox}
        \end{figure}
        \noindent
        This prompt explicitly considers the frequently occurring coarse-grained category (IV) Logical and Reasoning Errors and (I) Problem Understanding Errors.
        The reason for adopting this prompt is to determine whether more detailed error analysis leads to greater performance improvements.

        \paragraph{Dataset.}
        As a development split, we randomly sampled a total of 700 problems from the MATH test set, selecting 20 problems from each domain at each level.
        The phrase in the proposed prompt was determined based on the performance of Llama-3.3-70B-Instruct on the development split.
        The remaining 4,300 problems were used as a test split.

        \paragraph{Evaluation models.}
        We used three models as evaluation models:
        Llama-3.3-70B-Instruct, Qwen3-32B\footnote{https://huggingface.co/Qwen/Qwen3-32B}~\citep{qwen3}, and gemma-2-27b-it\footnote{https://huggingface.co/google/gemma-2-27b-it}~\citep{gemmateam2024gemma2improvingopen}.

       \paragraph{Evaluation metrics.}
        We set the temperature to $0.6$, and the results are averaged over 10 random seeds.
        The sequence length is set to 8,192, and we determine answer correctness by EM.
        To perform statistical tests between the proposed prompt and each baseline, we compute rank-biserial correlation (RBC) values and p-values using the Mann–Whitney $U$ test~\citep{mann1947test}.

    \subsection{Experimental Results} 
        To address RQ1, RQ2, and RQ3, we present our results in Table~\ref{tab:test_8192}.
        \begin{table}[t]
            \centering
            \resizebox{\linewidth}{!}{
            \begin{tabular}{l|r|r|r}
                \toprule
                 \multirow{2}{*}{}&\# incorrect ans. & \multicolumn{2}{c}{vs.\@ Ours} \\
                 \cmidrule(lr){2-2}
                 \cmidrule(lr){3-4}
                 & \multicolumn{1}{c|}{mean $\downarrow$ \ ($\pm$ std)}& \multicolumn{1}{c|}{RBC $\uparrow$} & \multicolumn{1}{c}{p-value $\downarrow$}\\
                \midrule
                \multicolumn{4}{c}{\textbf{Llama-3.3-70B-Instruct}} \\
                \midrule
                Default &1110.9 \ ($\pm$ 15.0) & 1.00 & 0.000 \\
                Pay Att. & 1104.8 \ ($\pm$ 10.4) & 1.00 & 0.000\\
                Coarse Att. & 1102.0 \ ($\pm$ 10.5) & 0.90 & 0.001 \\
                Ours  &1078.5 \ ($\pm$ \phantom{0}7.8) & -- \\
                \midrule
                \multicolumn{4}{c}{\textbf{Qwen3-32B}} \\
                \midrule
                Default &691.7 \ ($\pm$ 72.0) & 1.00 & 0.000\\
                Pay Att. &596.2 \ ($\pm$ 19.7) & 1.00 & 0.000\\
                Coarse Att. & 473.8 \ ($\pm$ \phantom{0}7.7) & 1.00 & 0.000 \\
                Ours &442.5 \ ($\pm$ \phantom{0}8.9)  & -- \\
                \midrule
                \multicolumn{4}{c}{\textbf{gemma-2-27b-it}} \\
                \midrule
                Default &2058.1 \ ($\pm$ 12.0) & 0.92 & 0.021 \\
                Pay Att. &2067.7 \ ($\pm$ 10.0) & 1.00 & 0.008\\
                Coarse Att. & 2067.0 \ ($\pm$ 19.6) & 0.92 & 0.001 \\
                Ours  &2021.6 \ ($\pm$ 15.5) & -- \\
                \bottomrule
            \end{tabular}
            }
            \caption{
            Results of Default Prompt, Pay Attention Prompt, Coarse Attention Prompt, and the proposed method. 
            \kosuke{The ``\# incorrect ans.'' column reports the mean and standard deviation of the number of incorrect answers over 10 runs.}
            \kosuke{The ``vs.\@ Ours'' column report}
            rank-biserial correlation values and p-values using the Mann–Whitney $U$ test.
            }
            \label{tab:test_8192}
        \end{table}

        \paragraph{RQ1: Is explicitly directing attention to considering reasoning errors effective?}
        As shown in Table~\ref{tab:test_8192}, the proposed prompt consistently outperformed both Default Prompt and Pay Attention Prompt across all evaluated models.
        In particular, the proposed prompt shows statistically significant improvements over Default Prompt across all evaluated models, with p-values < 0.05\@.
        This result demonstrates that classifying reasoning errors effectively improves reasoning ability.
        Furthermore, Pay Attention Prompt did not yield robust performance improvements over Default Prompt. 
        Specifically, for Llama-3.3-70B-Instruct, the mean scores of the two methods were within one standard deviation of each other. 
        For Qwen3-32B, their difference only slightly exceeded one standard deviation, whereas Pay Attention Prompt performed worse than Default Prompt on gemma-2-27b-it.
        
        These results suggest that merely encouraging attention is insufficient, and it is important to explicitly direct attention to error-prone aspects based on the classes of reasoning error.
        Figure~\ref{fig:ex_problem} presents an example of a reasoning error that the Pay Attention Prompt did not improve, but the proposed prompt corrected.
        In this example, Default Prompt and Pay Attention Prompt ignore the condition that your friend has at least 200 eggs, but the proposed prompt takes it into account 
        by confirming that all conditions in the text had been extracted.
        \begin{figure*}[t]
        \centering
        \begin{tcolorbox}
            \textbf{Problem:} 
            \colorbox{lightgray}{Your friend has an egg collection comprising at least \$$200$\$ eggs.} 
            ...(omitted)...\
            What is the minimum number of eggs that your friend could have?\\
            \textbf{Final answer contained in the dataset:} $313$\\
            \textbf{The output of Default and Pay Attention Prompts:}\\
            \#\# Step 1: Understand the problem\\
            The problem involves finding the minimum number of eggs a friend could have, given that when stored in dozen-egg containers, there is 1 egg left over, and when stored in baker-dozen-egg containers (holding 13 eggs), there is also 1 egg left over.\ ...(omitted)...\ \\
            Since \$N-1\$ must be a multiple of the LCM of 12 and 13, the smallest possible value for \$N-1\$ is 156. Therefore, the minimum number of eggs \$N\$ is \$156 + 1 = 157\$.
            The final answer is: \$\textbackslash\textbackslash boxed\{157\}\$\\
            \textbf{The output of proposed prompt:}\\
            \#\# Step 1: Understand the problem conditions\\
            \colorbox{lightgray}{The problem states that the friend has at least 200 eggs} and wants to store them in dozen-egg containers. \ ...(omitted)...\ \\
            \#\# Step 10: Verify the minimum solution\\
            \colorbox{lightgray}{Given that 157 is the first common solution but is less than 200}, and knowing that the pattern repeats every 156 eggs, the next possible solution would indeed be \$157 + 156 = 313\$. This satisfies both modular conditions and is above the minimum of 200 eggs. It's essential to verify that no smaller solution exists within the given constraints.
            The final answer is: \$\textbackslash\textbackslash boxed\{313\}\$
            \end{tcolorbox}
            \caption{
            An example of errors that did not improve with Pay Attention Prompt but improved with the proposed prompt.
            The output of Pay Attention Prompt in this example was identical to that of Default Prompt.
            \kosuke{The reasoning error in this example is classified into (I-2) Ignoring Conditions in the Problem Statement.}
            \colorbox{lightgray}{The text with a light gray background} refers to the condition that the number of eggs is at least 200.
            }
            \label{fig:ex_problem}
        \end{figure*}

        \paragraph{RQ2: Is fine-grained reasoning error classification effective?}
        As shown in Table~\ref{tab:test_8192}, the proposed prompt outperforms Coarse Attention Prompt across all evaluated models.
        Furthermore, similar to Pay Attention Prompt, Coarse Attention Prompt did not yield robust performance improvements.
        Specifically, Coarse Attention improved performance for Qwen3-32B but degraded it for gemma-2-27b-it.
        These results indicate that achieving robust performance improvements requires not only classifying errors into coarse classes but also classifying them into fine-grained reasoning error classes and explicitly highlighting specific aspects in which errors are likely to occur. 
        Therefore, detailed error analysis is important not only as a framework for evaluating models’ reasoning capabilities but also for designing effective prompts.

        \paragraph{RQ3: Are frequent reasoning errors model-specific? }
        As discussed above, not only Llama-3.3-70B-Instruct but also Qwen-3-32B and gemma-2-27b-it show performance improvements.
        This suggests that the reasoning errors analyzed in this paper are not specific to the Llama-3 family, but may instead reflect challenges common to LLMs of comparable scale.

\section{Conclusion}

    \kosuke{The mathematical reasoning ability of LLMs underlies a broad range of downstream applications and has the potential to advance scientific discovery.}
    We clarify challenges in current LLMs and offer insights into directions for improvement.
    
    We conducted a detailed qualitative analysis of the reasoning errors made by the LLM on the mathematical dataset.
    We first defined broad categories based on existing research, then further subdivided them into 21 finer-grained classes of reasoning errors.
    Our work clarified that current LLMs struggle to consider all the conditions in the problem statement and the prerequisites of the introduced theorems.


    Our results demonstrate that error analysis serves not only as a means of diagnosing a model's capabilities but also as a practical foundation for improving its performance.
    In particular, we show that the prompt that directs attention to frequently occurring errors can improve mathematical reasoning performance.
    These findings suggest that it is important to carefully evaluate model behavior and design prompts based on the evaluation results.

    We highlight two promising directions for future work:
    \begin{itemize}
        \item 
        \kosuke{Our manual reasoning error classification produced an annotated dataset consisting of problem statements, answers, reasoning processes, correctness labels, and reasoning error class labels. This dataset enables the training and evaluation of models for reasoning error classification and detection. Deploying such models is valuable not only for understanding model behavior, but also for applications such as constructing training data to mitigate errors~\citep{yu-etal-2025-self-error} and detecting hallucinations~\citep{manakul-etal-2023-selfcheckgpt}.}
        \item 
        \kosuke{
        Our fine-grained definition of the reasoning error classes enables LLMs used in the LLM-as-a-judge framework to assign scores across fine-grained aspects. 
        In particular, prior work has demonstrated that aspect-wise scoring is beneficial for LLM alignment~\citep{cui2024ultrafeedback}, and that  assigning rewards to the reasoning process itself improves reasoning capability~\citep{sun-etal-2025-efficient, ICLR2025_31a57804}. 
        Building on these findings, our fine-grained definition enables more precise evaluation, which can improve LLMs' reasoning capability through enhanced LLM-as-a-judge systems.}
    \end{itemize}


\section{Limitations}
    \kosuke{This paper is broadly divided into two parts: an analysis framework (\S\ref{sec:analysis}) and a method to reduce frequent errors, together with its evaluation (\S\ref{sec:method} and \S\ref{sec:experiment}). We discuss two limitations of this work below.}
    
    \kosuke{First, although our analysis in \S\ref{sec:analysis} provides manual error annotation for each LLM's response, our experiments in \S\ref{sec:experiment} report only aggregate accuracy and do not annotate the error class of each failed response, due to heavy annotation cost. Tracing individual responses and their error classes from Default Prompt to Proposed Prompt would provide further implications for understanding mathematical reasoning behaviors of LLMs, which we leave as a promising direction for future work.}
    
    \kosuke{Second, our experiments are limited to the MATH dataset and three open-weight models with parameter counts ranging from 27B to 70B, which is one of the most popular settings for evaluating LLMs' mathematical capabilities across a broader range of difficulty levels.
    Consequently, it remains unclear whether the observed improvements generalize to other mathematical benchmarks and models outside this size range.
    For example, smaller models have been reported to be more prone to calculation errors~\citep{seßler2024benchmarkinglargelanguagemodels}. However, we believe that our analysis framework would also provide valuable insights for examining LLMs' behavior in such settings.}%


\bibliography{ysdln0616, ysdln0616_arxiv}

\end{document}